\documentclass[12pt]{article}

\usepackage[a4paper,margin=1in]{geometry}
\usepackage[T1]{fontenc}
\usepackage[utf8]{inputenc}

\usepackage{amssymb}
\usepackage{amsmath}
\usepackage{booktabs}
\usepackage{multirow}
\usepackage{array}
\usepackage{graphicx}
\graphicspath{{figures/}{supplementary/figures/}{supplementary/}}
\usepackage{bm}
\usepackage{threeparttable}
\usepackage{rotating}
\usepackage[numbers,sort&compress]{natbib}
\usepackage{authblk}
\usepackage{hyperref}
\usepackage{fancyhdr}
\usepackage{fontspec}
\usepackage{setspace}
\usepackage{fontspec}
\usepackage{setspace}

\title{\bfseries NeoLoc-68: End-to-end 68-point neonatal facial landmark localisation in neonatal clinical environments}

\author[1]{Abdullah Bin-Obaid}
\author[2,3]{Maria M. Cobo}
\author[2]{Rebeccah Slater}
\author[1]{Lionel Tarassenko}
\author[1]{Mauricio Villarroel}

\affil[1]{Institute of Biomedical Engineering, Department of Engineering Science, University of Oxford, Oxford, United Kingdom}
\affil[2]{Department of Paediatrics, University of Oxford, Oxford, United Kingdom}
\affil[3]{Universidad San Francisco de Quito USFQ, Colegio de Ciencias Biologicas y Ambientales, Quito, Ecuador}

\date{}

\fancypagestyle{firstpage}{
  \fancyhf{}
  
  \fancyfoot[L]{\footnotesize\itshape This manuscript is a preprint and has not yet been peer reviewed. It is currently under consideration at Pattern Recognition Letters. The final published version, if accepted, may differ from this version.}
  \fancyfoot[R]{\thepage}
}

\begin{document}

\maketitle
\thispagestyle{firstpage}

\begin{center}
\small
\textbf{Corresponding authors:}\\
Abdullah Bin-Obaid, \href{mailto:abdullah.binobaid@eng.ox.ac.uk}{abdullah.binobaid@eng.ox.ac.uk}\\
Mauricio Villarroel, \href{mailto:mauricio.villarroel@eng.ox.ac.uk}{mauricio.villarroel@eng.ox.ac.uk}
\end{center}

\begin{abstract}
Facial landmark localisation is a prerequisite for developing automated, non-contact neonatal pain assessment methods. Clinicians use pain scales to judge the severity of pain, many of which rely on facial expression. However, facial landmark detectors trained on adult faces perform poorly in neonatal clinical environments due to frequent occlusions caused by medical equipment, varied head poses, and challenging imaging conditions, including motion blur triggered by sudden pain-related movements. We propose an end-to-end facial landmark detector capable of predicting 68 landmarks on neonatal faces in clinical environments. We combined 37,459 single-face images from 11 public datasets, standardised to 68-point markup, with 1,123 manually annotated frames from a neonatal research dataset (totalling over 76,000 landmarks). A YOLO-based keypoint model was adapted to regress the facial landmarks, initialised with weights from a pretrained neonatal face detector. On public datasets, our proposed model  achieved state-of-the-art performance: Normalised Mean Error (NME) = 5.37, Failure Rate (FR) = 12.5\%, Area Under the Cumulative Error Curve (AUC) at AUC\textsubscript{0.08} = 38.00\% and AUC\textsubscript{0.1} = 48.70\%. On the clinical neonatal test set, before fine-tuning, the model achieved the lowest Detection Failure Rate (DFR) = 5.3\% among all baselines and showed strong generalisation. After fine-tuning, performance improved further to NME = 6.36, FR = 22.30\%, DFR = 1.77\%, AUC\textsubscript{0.08} = 29.24\% and AUC\textsubscript{0.1} = 40.25\%. To the best of our knowledge, this represents the first end-to-end 68-point neonatal facial landmark detection model. With further dataset expansion and refinement, it could support downstream tasks in neonatal health monitoring and pain-related facial analysis.

\end{abstract}


\vspace{1em}

\setstretch{1.25}

\section{Introduction}

Facial landmark (or keypoint) detection is the task of localising key facial points such as the eyes, nose, and mouth. It has received considerable attention over the past decade due to its central role in various computer vision-based applications, such as facial expression analysis \cite{wu2019facial}. Most existing methods are trained for broad, non-specialised applications and are primarily based on adult or young children’s faces \cite{wan2022infanface,asgarian2019limitations}. Neonates in clinical environments, especially in the Neonatal Intensive Care Unit (NICU), present unique challenges, as they exhibit different anatomical proportions \cite{huelke1998overview} and are frequently partially occluded by medical equipment. In addition, they appear in unconstrained settings with varied lighting conditions and head poses.

Facial landmark detection in neonatal clinical environments has the potential to support a range of clinical applications, including automated pain assessment and objective monitoring of facial expressions associated with distress, discomfort, or behavioural state changes. However, the lack of annotated neonatal datasets, combined with the domain gap between adult and neonatal facial appearance, limits the direct applicability of existing models \cite{wan2022infanface,grooby2023neonatal}.

We developed an end-to-end facial landmark detection pipeline tailored to neonatal clinical environments, capable of localising 68 neonatal facial keypoints under challenging but realistic clinical conditions. Facial landmark detection in such environments enables various healthcare applications. Nevertheless, to the best of our knowledge, there is currently no model specifically trained to detect a large number of facial landmarks tailored to neonatal clinical settings.

\subsection{Related work}
Facial landmark detection has been addressed using various strategies that differ in the number and placement of predicted landmarks. Some approaches focus on as few as 5 landmarks, while others predict denser configurations, using 21, 68 or even up to 194 landmarks \cite{khabarlak2021fast, kazemi2014one}.

Dlib \cite{dlib09} is a widely used open-source C++/Python toolkit for face analysis.  It uses a Histogram of Oriented Gradients (HOG) based method for frontal face detection and a 68 point facial landmark detector implementing the method of Kazemi and Sullivan \cite{kazemi2014one}.

Face Alignment Network (FAN) \cite{bulat2017far} combined Stacked Hourglass Networks \cite{newell2016stacked} with a different bottleneck block from \cite{bulat2017binarized} to strengthen multi-scale feature aggregation. FAN also used a subnetwork based on ResNet to guide the conversion of 2D annotations to 3D. By doing this, the authors of FAN introduced a 3D landmarks dataset, combining augmented and synthetic images.

Pixel-in-Pixel Net (PIPNet) \cite{JLS21} proposed a regression head which uses low-resolution feature maps, removing the need for the upsampling layers for the heatmap regression while preserving accuracy and reducing computation. The authors of PIPNet also introduced a Neighbor Regression Module (NRM) that predicts nearby landmarks and fuses them at inference to enforce local shape consistency.

Shape Preserving with Graph Attention (SPIGA) \cite{SPIGA_model} combined heatmaps from different stages of the stacked Hourglass backbone with a cascade of attention regressors to learn geometric relationships between landmarks. SPIGA also estimates the head pose, which helps initialise landmark localisation. These additions aim to improve facial shape consistency and robustness to occlusion.

High-Resolution Network (HRNet) \cite{wang2020deep} maintained high-resolution representations end-to-end by running parallel branches of multi-resolution and repeatedly fusing them, rather than using a downsample-upsample design. The authors proposed several models used for keypoint estimation, semantic segmentation and detection.

Wan et al. \cite{wan2022infanface} introduced the Infant Annotated Faces (InfAnFace) dataset and trained HRNet-based models. They curated 410 web-sourced (Google/YouTube) infant images annotated with 68 landmarks and pose attributes, and released the annotations and code. These steps aimed to bridge the infant-adult landmarking gap, improving accuracy via joint training and fine-tuning with wide-rotation augmentation. However, the images were collected from the internet and depict infants in non-clinical, everyday environments (e.g., at home or outdoors), rather than neonates in clinical settings with medical equipment, occlusions, and hospital imaging conditions.

Grooby \cite{grooby2023neonatal} retrained YOLO-based detectors for neonatal face and 6 landmark localisation. The model was trained on a dataset of 455 manually annotated images, split into 366 for training and 89 for testing. These images were derived from three neonatal datasets including  Newborn Baby Heart Rate estimation database (NBHR)~\cite{huang2021neonatal}, infant Classification Of Pain Expressions videos (iCOPEvid)~\cite{brahnam2023neonatal}, and InfAnFace~\cite{wan2022infanface}. Landmark performance was reported using normalised mean error computed with the square root of the face bounding-box area rather than inter-ocular distance.

\section{Materials and methods}

\subsection{Datasets}
We used 11 publicly available facial landmark datasets and an additional neonatal research dataset. Datasets were selected if their annotations were compatible with or convertible to the Intelligent Behaviour Understanding Group (iBUG) 68-point landmark format \cite{sagonas2013300}, allowing all data sources to be standardised to a unified landmark configuration for consistent training and evaluation.

\noindent
\textbf{Extended Multi-Modal Verification for Teleservices and Security (XM2VTS)} \cite{messer1999xm2vtsdb} comprises 2,360 facial images of 295 adult individuals, captured in a controlled studio environment. Each subject was recorded over four sessions, each spaced approximately one month apart, to capture temporal variability in facial appearance, resulting in 8 images per individual.

\noindent
\textbf{Multi-session Pose, Illumination, and Expression (MultiPIE)} \cite{gross2010multi} was captured in a controlled laboratory environment with systematic variations in pose, illumination, and expression. It includes 318 unique individuals, with image counts ranging from 4 to 36 per individual (median = 22.5). A total of 6,665 frontal images are annotated with 75 landmarks.

\noindent
\textbf{Face Recognition Grand Challenge (FRGC)} \cite{phillips2005overview} contains 4,950 facial images captured in a controlled laboratory environment with consistent lighting and background conditions. It includes 557 unique individuals, with between 1 and 30 images per individual (median = 7). Each image is annotated with 75 landmarks.

\noindent
\textbf{Labeled Face Parts in the Wild (LFPW)} \cite{belhumeur2013localizing} includes 1,035 facial images collected from Google Images, Flickr, and Yahoo Images. While originally annotated with 29 landmarks, the iBUG group later provided 68-point annotations for this dataset. Some images contain multiple visible faces, and the number of annotated faces per image may vary. The identity of each individual is not specified.

\noindent
\textbf{HELEN dataset} \cite{Le2012InteractiveFF} comprises 2,330 images collected from Flickr, representing a variety of poses, lighting conditions, and facial expressions. It is one of the most widely used datasets for landmark localisation tasks, with 68-point annotations provided by the iBUG group. Some images contain multiple visible faces, and the number of annotated faces per image may vary. The identity of each individual is not specified.

\noindent
\textbf{Annotated Faces in the Wild (AFW)} \cite{Zhu2012FaceDP} includes 337 facial images. The 68 landmarks annotation for this dataset is provided by the iBUG group. Some images contain multiple visible faces, and the number of annotated faces per image may vary. The identity of each individual is not specified.

\noindent
\textbf{300-W} \cite{sagonas2016300} consists of 600 facial images annotated with 68 landmarks. It was originally released as part of the 2013 Automatic Facial Landmark Detection in-the-Wild Challenge \cite{sagonas2013300}. It has since become a benchmark for facial landmark localisation. The dataset includes two categories,  indoor and outdoor images, covering a range of facial expressions, lighting conditions, and occlusions. Some images contain multiple visible faces, and the number of annotated faces per image may vary. The identity of each individual is not specified.

\noindent
\textbf{iBUG} \cite{sagonas2016300} consists of 135 images annotated with 68 landmarks and originally served as the challenging test subset of the 300-W benchmark. It contains images with increased variation in pose, expression, and occlusion. Some images contain multiple visible faces, and the number of annotated faces per image may vary. The identity of each individual is not specified.

\noindent
\textbf{Caltech Occluded Faces in the Wild (COFW)} \cite{burgos2013robust} includes 507 images, each annotated with 75 landmarks and varying levels of facial occlusion. Some images contain multiple visible faces, and the number of annotated faces per image may vary. The identity of each individual is not specified.

\noindent
\textbf{Menpo2D} \cite{zafeiriou2017menpo} consists of 11,989 images, each annotated with 75 landmarks. The dataset includes a diverse range of facial appearances, including variation in pose, lighting, and expression. Some images contain multiple visible faces, and the number of annotated faces per image may vary. The identity of each individual is not specified.

\noindent
\textbf{Wider Facial Landmarks in the Wild (WFLW)} \cite{wayne2018lab} combines 6,551 images available for research. The dataset exhibits a broad range of variations in pose, occlusion, and illumination across 60 categories, with each category containing between 19 and 439 images. All images are annotated with 98 landmarks.

\textbf{The neonatal research dataset} comprises 226 video recordings (113 recording sessions involving 113 independent infants). Each recording session included two 45-second videos, one recording of the clinically-required heel lance procedure and one paired control recording involving a non-painful stimulus.

These recordings were collected during the Petal trial \cite{cobo2022multicentre,hauck2024effect} (approved by the London-South East Research Ethics Committee, reference 21/LO/0523) and the ongoing study ‘Investigating Pain in the Developing Human Brain’ (South Central - Oxford C Research Ethics Committee, reference 12/SC/0447). Written informed parental consent was obtained for all neonates prior to participating, and the studies conformed to the Declaration of Helsinki and Good Clinical Practice standards.

Data were collected between 2012 and 2023 at the John Radcliffe Hospital (Oxford University Hospitals NHS Foundation Trust, UK) and the Royal Devon and Exeter Hospital (Royal Devon University Healthcare NHS Foundation Trust, UK). Postmenstrual age of the neonates ranged from 33 to 43 weeks, with a mean $\pm$ SD of 39.3 $\pm$ 1.9 weeks. All infants underwent clinically-required heel lance procedures, which are routinely used to collect blood samples from an infant’s foot for diagnostic testing. The studies took place on the postnatal wards and neonatal units (including the Low Dependency Unit (LDU), High Dependency Unit (HDU), and Intensive Therapy Unit (ITU)/NICU. Data collected in these studies included hand-held video recordings to capture behavioural responses and facial expressions to enable pain scoring using the Premature Infant Pain Profile-Revised (PIPP-R) scale \cite{stevens2014premature}.

\subsection{Datasets preprocessing} \label{sec:data_preprocessing}
Some images in the selected datasets contained multiple visible faces, while landmark annotations were not always provided for every visible face in the image. This can negatively impact model training, as the detector may predict landmarks for unannotated faces, which would be treated as false positives.

We applied four preprocessing stages: (1) single-face image selection and identification of multi-face cases; (2) standardising all landmark annotations to the 68-point iBUG format; (3) generating bounding boxes from landmark coordinates; and (4) cropping multi-face images to isolate the annotated face.

By following this pipeline, we produced single-face images with corresponding ground truth bounding boxes and landmark annotations, suitable for training and testing face detection and landmark localisation models.

\subsubsection{Single face image selection} 
To ensure that each training image contained only one annotated face, we employed three state-of-the-art face detection models: SCRFD \cite{guo2021sample}, DSFD \cite{li2019dsfd}, and MTCNN \cite{zhang2016joint} as independent detectors to screen the datasets.

A single-face image was defined as one in which all three detectors identified exactly one face, and the predicted bounding boxes overlapped in the same region of the image, confirmed by an Intersection-over-Union (IoU) of at least 0.6 between all detections. Images satisfying this condition were treated as single-face images and used directly.

Images in which multiple faces were detected, or where detector agreement was inconsistent, were not excluded from the dataset. Instead, they were forwarded to the cropping stage to isolate the annotated face.

\subsubsection{Standardising landmark annotation in public datasets}
\begin{figure}[!htbp]
   \centering
   \includegraphics[width=\textwidth]{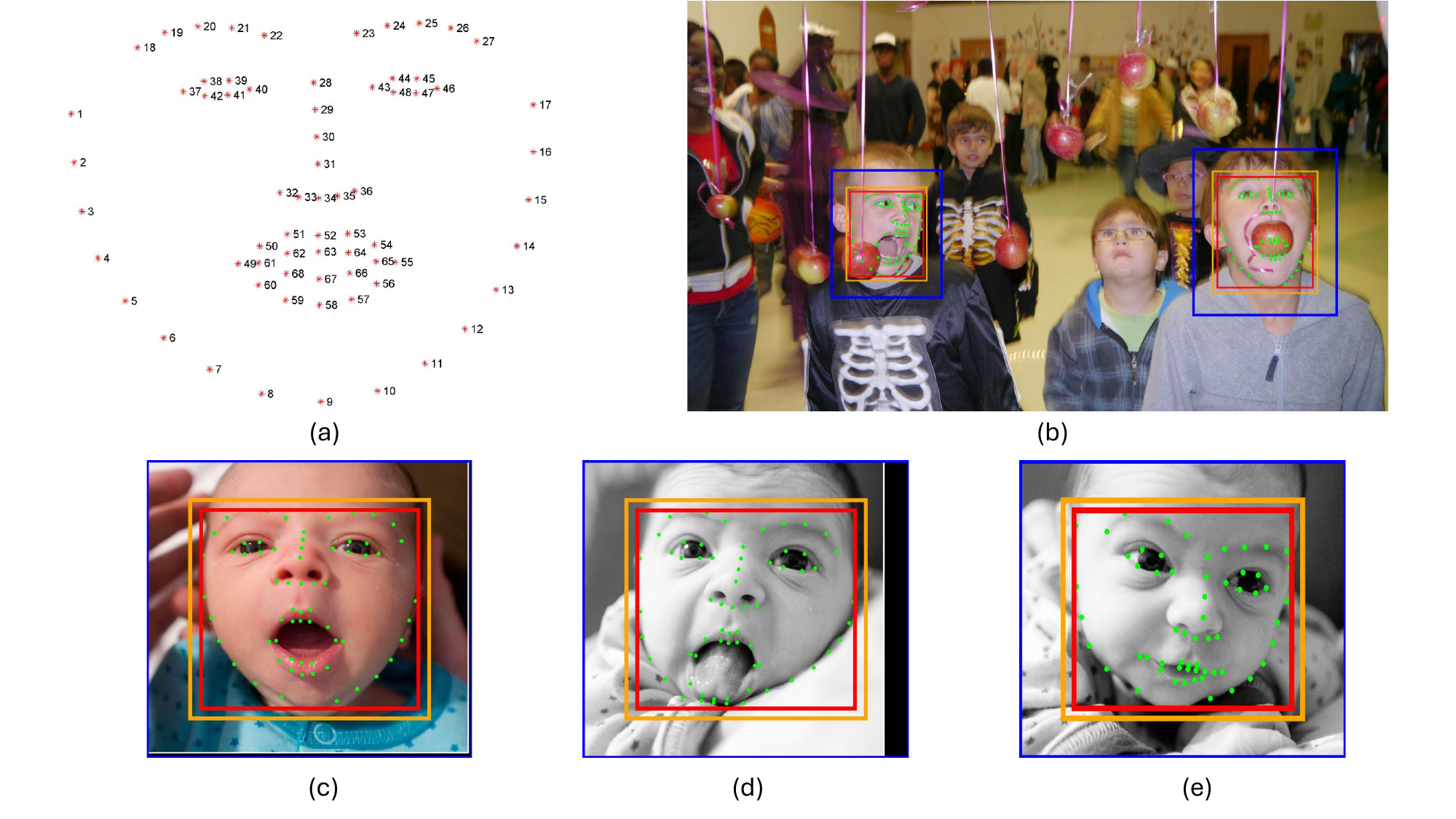}
   \caption
   {
   \textbf{The 68-point facial landmark annotation scheme and example annotations.} The 68-point facial landmark configuration and an example annotation. (a) The landmarks are numbered in order and span the key facial regions, including the jawline, eyebrows, eyes, nose, and mouth (image adapted from \cite{sagonas2013300}). (b) Example image showing landmark annotations (green), the landmark-derived bounding box (red), the expanded ground-truth training bounding box (orange), and the expanded cropping region (blue). Multiple faces are visible in the image, but only two are annotated with facial landmarks. (c-e) Additional examples of cropped face regions from the publicly available 300-W dataset, showing the corresponding landmark annotations and bounding boxes. Image (b) has been slightly cropped to improve visual alignment, and images (c-e) are from publicly available 300-W dataset.
   }
   \label{fig:68_markup}
\end{figure}

The selected datasets vary in keypoint count and annotation format, requiring standardisation to the 68-landmarks markup. This widely used scheme, shown in Fig~\ref{fig:68_markup}(a), was originally introduced by Gross et al. \cite{gross2010multi} in the MultiPIE dataset and later adopted in the 300-W dataset as part of the 300 Faces In-The-Wild Challenge \cite{sagonas2013300}.

For the XM2VTS, MultiPIE, FRGC, COFW and Menpo2D datasets, we removed non-standard points such as bounding box corner points, duplicate nose tip and mouth edge points, and eye pupil annotations. For MultiPIE and Menpo2D, we included only the semi-frontal images, as the profile images contained fewer than 68 landmarks.

The WFLW dataset closely follows the iBUG markup but with three notable differences: (1) For the face edges (the jawline), WFLW includes 33 points, essentially inserting an additional point between each pair of the 17 iBUG defined points. We resolved this by retaining every other point. (2) For the eyebrows, WFLW provides 9 points per brow (five along the upper edge and four below it), whereas iBUG specifies five points per brow. We removed the four additional lower eyebrow points to align with the iBUG format. (3) Finally, we removed the two eye pupil points and, for each eye, the two eyelid points roughly above and below the pupil, resulting in a dataset of 68 landmarks consistent with the iBUG markup. Examples of the original annotations and their corresponding standardised 68 landmarks conversions are provided in Supplementary Material S1.

\subsubsection{Generating face bounding boxes}

Not all selected datasets provided ground-truth face bounding box annotations. As the detection component of our model requires bounding boxes for training, we generated a bounding box for each image directly from the 68 landmark coordinates.

For each image, we computed the smallest rectangular box enclosing all landmark points, hereafter referred to as the landmark-derived bounding box. To provide a small margin around the face, this box was expanded by a factor of 1.1. The resulting box was used as the ground-truth face bounding box during training. The orange box in Fig~\ref{fig:68_markup}(b) illustrates the expanded landmark-derived bounding box used as the ground-truth face box during training.

\subsubsection{Cropping multi-face images}

For images containing multiple visible faces, we cropped the image around the landmark-derived bounding box corresponding to the annotated face, expanded by a factor of 1.5. This expansion preserves surrounding facial context during cropping. If the expanded region extended beyond the image boundaries, zero-padding was applied to maintain a valid crop. The blue box in Fig~\ref{fig:68_markup}(b) illustrates the cropping region.

\subsubsection{Preprocessing the neonatal research dataset} 
We extracted a total of 1,130 frames from the neonatal research video recordings, with 10 frames per recording session. From each recording session,  we extracted nine frames from the heel lance (acute procedural pain) video and one frame from the control (non-painful stimulus) video. Infants are typically less calm and display more facial expressions in the heel lance videos compared to the control ones. Frames in which the face was fully occluded, for example by clinicians’ hands, medical equipment, or infant self-movement, were excluded because the facial landmarks could not be reliably annotated. After exclusion, 1,123 images remained.

Each image required annotation with 68 facial landmarks following the iBUG format, resulting in 76,364 landmarks to be labelled. The data were manually annotated by one of the authors using Computer Vision Annotation Tool (CVAT) software~\cite{CVAT}.

\subsubsection{Dataset summary} 
\begin{figure}[!htbp]
	\centering
	\includegraphics[width=\textwidth]{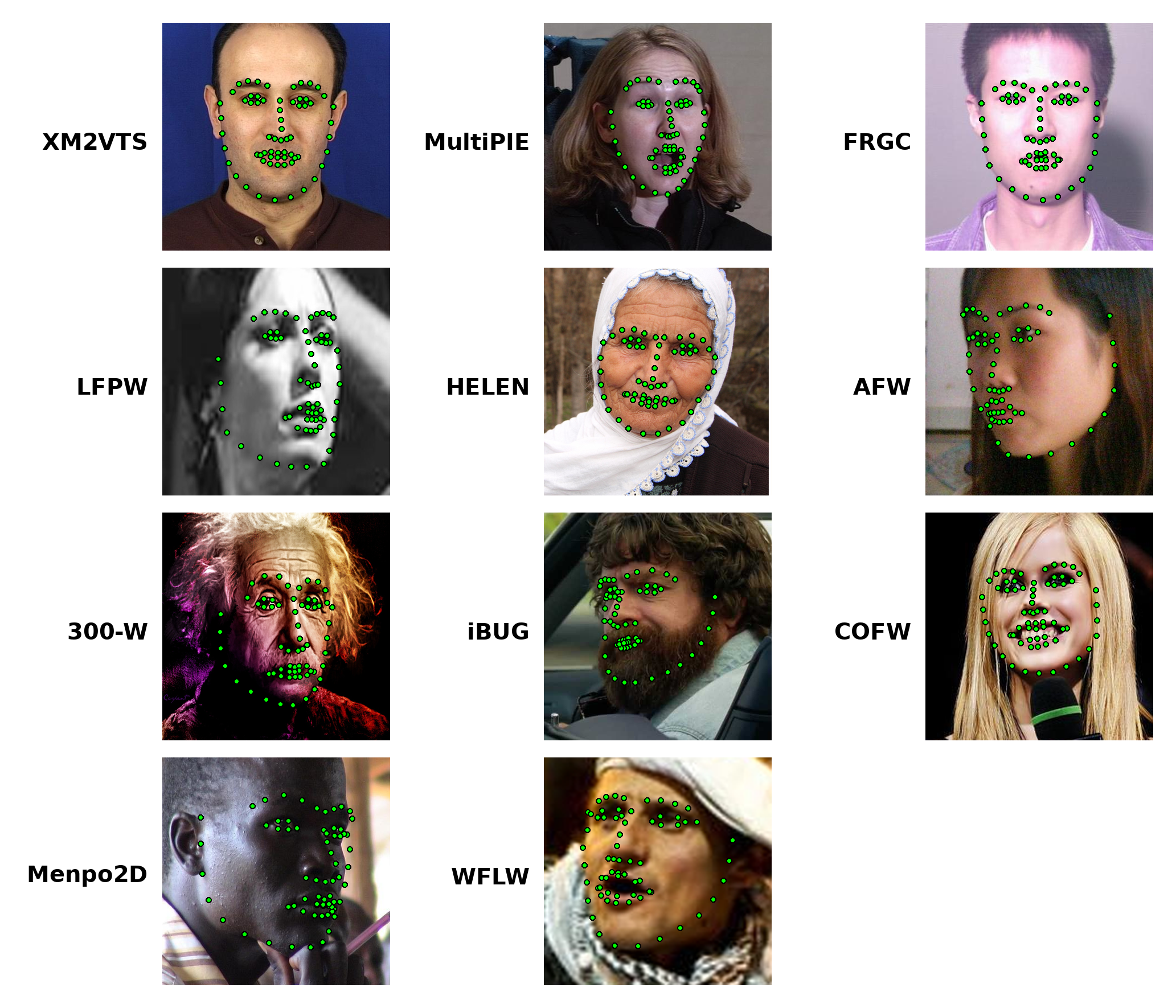}
	\caption
	{
		\textbf{Example images from the selected facial landmark datasets used in this work.} Each image is accompanied by its dataset name on the left and is shown with its corresponding set of 68 annotated facial landmarks, reformatted where necessary to follow the iBUG markup scheme. Images have been cropped around the face region to provide consistent framing and improve visual comparability across datasets.
	}
	\label{fig:examples_public_datasets}
\end{figure}

As summarised in Table \ref{tab:keypoints_dataset_summary}, our study employs 11 public facial landmark datasets totalling 37,459 images across training and test splits. To bridge the domain gap to our intended clinical setting, we additionally include the neonatal research dataset for domain adaptation. Example images from the public sets, with standardised 68 point annotations following the iBUG markup, are shown in Fig \ref{fig:examples_public_datasets}.

\begin{table}[htbp]
\centering
\caption{Summary of the facial landmark datasets used.}
\label{tab:keypoints_dataset_summary}
\begin{tabular}{@{}lrrr@{}}
\toprule
\textbf{Dataset} & \textbf{Total images} & \textbf{Training images} & \textbf{Test images} \\
\midrule
\midrule

\multicolumn{4}{@{}l}{\textit{\textbf{Public datasets}}} \\
\midrule
\hspace{4mm}XM2VTS \cite{messer1999xm2vtsdb}       & 2,360  & 1,888  & 472   \\
\hspace{4mm}MultiPIE \cite{gross2010multi}         & 6,665  & 5,348  & 1,317 \\
\hspace{4mm}FRGC \cite{phillips2005overview}       & 4,950  & 3,980  & 970   \\
\hspace{4mm}LFPW \cite{belhumeur2013localizing}    & 1,035  & 835    & 200   \\
\hspace{4mm}HELEN \cite{Le2012InteractiveFF}       & 2,330  & 1,864  & 466   \\
\hspace{4mm}AFW \cite{Zhu2012FaceDP}               & 337    & 269    & 68    \\
\hspace{4mm}300-W \cite{sagonas2016300}            & 600    & 480    & 120   \\
\hspace{4mm}iBUG \cite{sagonas2016300}             & 135    & 108    & 27    \\
\hspace{4mm}COFW \cite{burgos2013robust}           & 507    & 405    & 102   \\
\hspace{4mm}Menpo2D \cite{zafeiriou2017menpo}      & 11,989 & 9,591  & 2,398 \\
\hspace{4mm}WFLW \cite{wayne2018lab}               & 6,551  & 5,455  & 1,096 \\
\midrule
\textbf{Public datasets total}         & \textbf{37,459} & \textbf{30,223} & \textbf{7,236} \\
\midrule
\multicolumn{4}{@{}l}{\textit{\textbf{Clinical dataset}}} \\
\midrule
\hspace{4mm}Neonatal research \cite{cobo2022multicentre,hauck2024effect} & 1,123 & 557 & 566 \\
\midrule
\textbf{Totals}                        & \textbf{38,582} & \textbf{30,780} & \textbf{7,802} \\
\bottomrule
\end{tabular}
\end{table}

\subsection{Model architecture}
Although a variety of architectures have been proposed in the literature specifically for facial landmark detection, the underlying task is simply the localisation of points of interest. Consequently, a given landmarks detector, independent of its original application domain, can be adapted to facial landmark detection through task-specific modifications.

We employ the YOLO-POSE model \cite{YOLO_2023} (hereafter referred to as the Keypoints model) for keypoints detection. This model is a single-stage model and follows a top-down approach: it jointly predicts the object of interest (in this case, a face) and its associated keypoints. The original YOLO-POSE model was designed for human pose estimation with 17 keypoints. To adapt it for facial landmark detection, we modify the final regression head to output 68 keypoints represented by its $(x, y)$ coordinates in the image.

The primary motivation for selecting this YOLO model is its ability to leverage pre-trained weights from a large-scale face detection model trained on millions of images (see Supplementary Material S2), providing a strong prior for learning facial structure.

\subsection{Model training}  \label{sec:model_training} 
We trained NeoLoc-68 using weight transfer initialisation together with task-specific data augmentation to improve generalisation. To begin the data splitting process, we identified two types of datasets, each requiring a different strategy. The first type consists of datasets that originally contained multiple faces per image or lacked participant identifiers, making stratified splitting infeasible. Instead, we performed a simple 80\%-20\% split for training and testing, respectively. This approach was used for the LFPW, HELEN, AFW, iBUG, COFW and Menpo2D datasets.

The second type includes datasets collected in controlled environments, where each image contains a single face associated with a participant ID or categorical metadata suitable for stratification. We examined the available stratification metadata together with the number of images associated with each group. For the 300-W and WFLW datasets, we stratified the split using the provided category labels and applied an 80\%-20\% training-testing split within each category to maintain a consistent distribution.

For datasets containing participant identity labels and image counts per participant, we accounted for the unbalanced distribution of images while preventing subject leakage across splits. We binned the number of images per participant into four quartiles based on the overall distribution, then used these bins to perform a stratified split at the participant level. This ensured that both the training and testing sets included a balanced mix of participants contributing varying amounts of data. While this method slightly affected the number of images in each split, because participants contributed different numbers of images, the total approximated an 80\%-20\% split, with the proportion maintained at the participant level.

We split the training dataset into 80\% training and 20\% validation sets using 5-fold cross-validation. We maintained the same stratification strategy described earlier to ensure consistency and to prevent any data leakage between training and validation sets.

For the neonatal research dataset, we split the dataset into approximately 50\% for fine-tuning and 50\% for testing, ensuring that the split was performed at the participant level to prevent data leakage. Specifically, no participant appeared in both the fine-tuning and testing sets. We then further divided the fine-tuning set into 80\% for training and 20\% for validation.

\subsection{Weight initialisation}
\label{sec:Weight_initialisation}

\begin{figure}[!htbp]
	\centering
	\includegraphics[width=\textwidth]{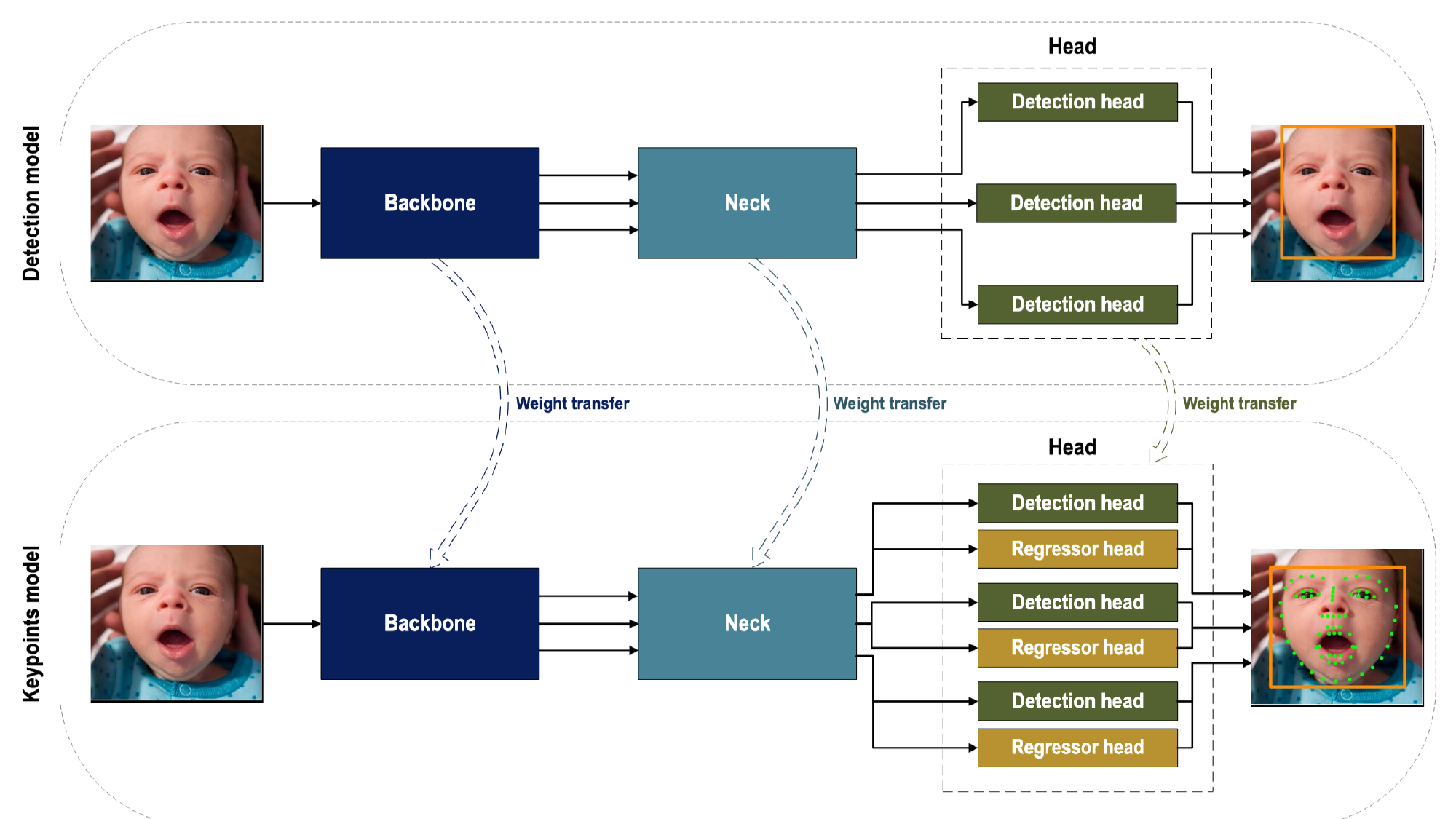}
	\caption
	{
		\textbf{Illustration of the weight transfer process from the detection model to the keypoints model.} The high-level architectures of both models: the face detection model (top) and the keypoints model (bottom). Both share the same backbone (dark blue rectangle), neck (light blue rectangle), and detection heads (green rectangles). The keypoints model includes additional regressor heads (gold rectangles) for keypoints localisation.  Infant image from the publicly available 300-W dataset.
	}
	\label{fig:transfer_weights}
\end{figure}

We initialised the backbone, neck, and detection heads using pre-trained weights from our face detection model. Fig~\ref{fig:transfer_weights} illustrates the overall weight transfer process. Details of the face detection model are provided in Supplementary Material S2.

This initialisation strategy leverages representations learned from large-scale face detection training. NeoLoc-68 itself operates as an end-to-end model that performs landmark localisation directly without requiring a separate face detector during inference.

\subsection{Data augmentation} \label{sec:data_augmentation} 

We adopted a standard set of data augmentation techniques widely used in facial landmark detection. We applied random rotation within $[-30^\circ, +30^\circ]$, consistent with common augmentation ranges used in prior facial landmark studies \cite{guo2019pfld,feng2018wing}, to capture moderate head pose variations. In addition, horizontal flipping with a probability of 0.5 and scaling with a factor randomly sampled from the range $[1 \pm 0.25]$ were applied to help the model generalise to different face sizes and distances from the camera \cite{li2020structured,JLS21}. When applying horizontal flips, we ensured that the x-coordinates of all landmarks were adjusted accordingly, and that symmetrical landmark indices were swapped. For example, the left and right eye corners were swapped to preserve the correct anatomical correspondence in the 68-point format. We did not include vertical flipping, as upside-down faces are not representative of neonates in clinical environments and would introduce unrealistic variations that are unlikely to occur in the target application domain. To account for slight positional shifts and partial occlusions, we applied random translation with a maximum shift of 10\% of the image dimensions in both horizontal and vertical directions. We set the shear factor to randomly range within $[-25^\circ, +25^\circ]$, and computed the Hue Saturation Value (HSV) gains to be 0.57, 0.53, and 0.43 for H, S and V, respectively (see Supplementary Material S3 for further details on HSV selection).

\subsection{Loss function} 

The total loss function, $\displaystyle \bm{\mathcal{L}}_{\mathrm{total}}$, that we used to train our model is defined as the sum of the detection loss and the keypoints loss, expressed as:

\begin{equation}
	\bm{\mathcal{L}}_{\mathrm{total}} = \mathcal{L}_{\mathrm{dt}} + \mathcal{L}_{\mathrm{keypoints}}
\end{equation}

\noindent
where $\mathcal{L}_{\mathrm{dt}}$ is the detection loss expressed as:

\begin{equation}
    \mathcal{L}_{\mathrm{dt}}
    = \lambda_{\mathrm{box}} \,\mathcal{L}_{\mathrm{iou}}
    + \lambda_{\mathrm{cls}} \,\mathcal{L}_{\mathrm{cls}}
    + \lambda_{\mathrm{dfl}} \,\mathcal{L}_{\mathrm{dfl}}
\end{equation}

\noindent
The detection loss $\mathcal{L}_{\mathrm{dt}}$ is described in detail in Supplementary Material S4.

The keypoints loss $\mathcal{L}_{\mathrm{keypoints}}$ is a custom loss introduced by Ultralytics \cite{YOLO_2023} adapted from Object Keypoint Similarity (OKS) formulation that is used in Common Objects in Context (COCO) keypoint evaluation. It accounts for the spatial accuracy of predicted keypoints, their visibility, and object scale, defined as:

\begin{equation}
	\begin{aligned}
		\mathcal{L}_{\mathrm{keypoints}}
		&=
		\lambda_{\mathrm{kpt}} \,\mathcal{L}_{\mathrm{kpt}}
		\;+\;
		\lambda_{\mathrm{kobj}} \,\mathcal{L}_{\mathrm{kobj}}
	\end{aligned}
	\label{eq:keypoint_loss}
\end{equation}

\noindent
$\mathcal{L}_{\mathrm{kobj}}$ is the keypoint visibility (``objectness'') loss and $\lambda_{\mathrm{kobj}}$ is its gain. As we are training our model in 2D, i.e., the datasets we have contain only $(x,y)$ coordinates without visibility flag, then $\mathcal{L}_{\mathrm{kobj}}$ is not used. Every landmark is assumed visible; therefore, this term is always zero. The term $\mathcal{L}_{\mathrm{kpt}}$ is the primary landmark localisation loss in our implementation and $\lambda_{\mathrm{kpt}}$ is its gain. $\mathcal{L}_{\mathrm{kpt}}$ is defined as:

\begin{equation}
	\begin{aligned}
		\mathcal{L}_{\text{kpt}} = \frac{1}{\sum_k m_k}  \sum_k (1 - \exp(-e_k)) m_k
	\end{aligned}
	\label{eq:OKS}
\end{equation}

\noindent
where $m_k \in \{0, 1\}$ serves as a mask to exclude absent landmarks, and $e_k$ is the normalised squared error for the $k^{th}$ keypoint, expressed as:

\begin{equation}
	\begin{aligned} \label{eq:normalised squared}
		e_k = \frac{d_k}{8 \sigma_k^2 (A+\varepsilon)}
	\end{aligned}
\end{equation}

\noindent
where $\sigma$ is the OKS prior $\sigma_k=\frac{1}{K}$ with $K=68$ as from the 68 facial keypoints, $A$ is the ground truth face box area and $\varepsilon$ is a small constant $\varepsilon =10^{-9}$ added to the denominator to avoid dividing by zero. The smaller the $e_k$ the more accurate the keypoints prediction is.

The $d_k$ term in Eq~(\ref{eq:normalised squared}) is the squared Euclidean distance error between the predicted points $(\hat{x},\hat{y})$ and the ground truth points $(x,y)$ and $k \in \{1, \dots, K\}$ with $K=68$. $d_k$ is expressed as:

\begin{equation}
	\begin{aligned}
		d_k = (\hat{x}_k - x_k)^2 + (\hat{y}_k - y_k)^2
	\end{aligned}
\end{equation}

\subsection{Validation monitoring metrics} 
During training, we monitored the training and validation keypoints loss $\mathcal{L}_{\mathrm{kpt}}$, as well as the officially implemented COCO-style OKS, which is computed after each epoch to assess keypoint localisation accuracy. The OKS metric shares the same core exponential structure as the loss in Eq~(\ref{eq:OKS}), but is instead used to compute a similarity score between the predicted and ground truth keypoints. OKS is defined as:

\begin{equation}
	\begin{aligned}
		\text{OKS} = \frac{\sum_{k=1}^{K} \exp\left(-e_k \right) \cdot \delta(v_k > 0)}{\sum_{k=1}^{K} \delta(v_k > 0)}
	\end{aligned}
\end{equation}

\noindent
where $K$ is the number of keypoints, $e_k$ is defined as in Eq~(\ref{eq:normalised squared}), $v_k \in \{0, 1, 2\}$ is the visibility flag for keypoint $k$, and $\delta(v_k > 0)$ is an indicator function equal to 1 if the keypoint is annotated, and 0 otherwise. This ensures that only labelled keypoints contribute to the calculation of OKS. The OKS score ranges from 0 to 1, with higher values indicating better agreement between prediction and ground truth.

To evaluate the model's performance during training, alongside the keypoints loss, we used two metrics derived from the OKS: the Average Precision (AP) at a fixed threshold of 0.5 ($\text{AP}_{50}$), and the mean AP averaged over multiple OKS thresholds ranging from 0.5 to 0.95 in steps of 0.05 ($\text{mAP}_{50:95}$).

\subsection{Training procedures} \label{sec:implementation_method}

We initialised training using the pre-trained weights from the face detection model. Following convention, we adopted the default YOLO input resolution of $640 \times 640$ pixels.

We modified the weight factor for the keypoint loss gain ($\lambda_{\mathrm{kpt}}$ in Eq~(\ref{eq:keypoint_loss})) by increasing it by 25\%. This adjustment aims to penalise the model more heavily when it fails to accurately localise keypoints. The rationale behind this modification stems from the increased complexity of the task; whereas the model previously focused on full body keypoints, it is now required to predict facial landmarks. Additionally, the number of keypoints has increased from 17 to 68, further justifying the need for a stronger emphasis on keypoint accuracy.

We trained the model for up to 100 epochs with a batch size of 32. During training, we monitored the keypoint loss and evaluation metrics ($\text{AP}_{50}$ and $\text{mAP}_{50:95}$). Early stopping was applied based on validation performance to prevent overfitting, and the final model was selected as the checkpoint achieving the highest validation mAP.

We optimised the model using the AdamW optimiser (initial learning rate = 0.002, $\beta_1$ = 0.90, $\beta_2$ = 0.999, weight decay = 0.001), with a cosine learning rate schedule reducing the learning rate from 0.002 to 0.0002 during training.

\subsection{Fine-tuning on the neonatal research dataset} 
We fine-tuned the NeoLoc-68 model that we trained on public datasets using the neonatal research data. As the task remains unchanged and only the domain has shifted, we kept the data augmentation pipeline described in section \ref{sec:data_augmentation}, updating only the HSV parameters to match the clinical neonatal images. The updated HSV values, computed from the neonatal research dataset, are 0.41 for Hue, 0.24 for Saturation, and 0.31 for Value.

We fine-tuned the model for 100 epochs with the backbone and neck layers frozen, with the batch size reduced to 16. We continued to use the AdamW optimiser with an initial learning rate of 0.0005, $\beta_{1}=0.90$, $\beta_{2}=0.999$, and weight decay of 0.01 to mitigate over-fitting. We adopted the same strategy for monitoring model performance and selecting the best-performing epoch as described in section \ref{sec:implementation_method} above.

\subsection{Evaluation metrics}

We evaluated model performance using Normalised Mean Error (NME), Failure Rate (FR), Detection Failure Rate (DFR), and the Area Under the Cumulative Error Distribution Curve (AUC). We computed NME between the predicted 68 facial landmarks and the ground-truth annotations as:

\begin{equation}
\label{eq:NME}
\mathrm{NME}_i =
\frac{1}{K}
\sum_{k=1}^{K}
\frac{\left\| \hat{x}_{ik} - x_{ik} \right\|_2}{d_i}
\end{equation}

\noindent
where $K$ is the number of facial landmarks, $\hat{x}_{ik}$ and $x_{ik}$ denote the predicted and ground-truth coordinates of the $k^{\text{th}}$ landmark, respectively, and $d_i$ is a normalisation factor. We defined $d_i$ as the inter-ocular distance, computed as the Euclidean distance between the eye centres obtained from the mean of the six landmarks of each eye:

\begin{equation}
d_i =
\left\|
\frac{1}{6}\sum_{j=36}^{41}x_{ij}
-
\frac{1}{6}\sum_{j=42}^{47}x_{ij}
\right\|_2
\end{equation}

\noindent
where $x_{ij} \in \mathbb{R}^2$ denotes the ground-truth 2D coordinates of the $j^{\text{th}}$ facial landmark in image $i$.

Each test image contained a single labelled face; however, detectors may still return multiple candidate faces, including false positives. We therefore included only retained landmark predictions from successful detections when computing NME, and excluded images with no valid detections. We regarded a prediction as failed when $\mathrm{NME} > 0.08$, and calculated FR as:

\begin{equation}
\mathrm{FR} =
\frac{F}{N}\times100\%
\end{equation}

\noindent
where $N$ denotes the number of test images in which a face was successfully detected and $F$ is the number of failed predictions.

We calculated DFR when the model failed to return 68 landmarks for the face of interest, either because no face was detected or because the best IoU with the ground-truth box was below 0.50:

\begin{equation}
\mathrm{DFR} =
\frac{N_{\text{fail}}}{N_{\text{total}}}.
\end{equation}

We computed the cumulative error distribution (CED), which shows the proportion of images whose NME does not exceed a selected threshold. For a given error threshold $\tau$:

\begin{equation}
\mathrm{CED}(\tau) =
\frac{1}{N}
\sum_{i=1}^{N}
1(\mathrm{NME}_i \le \tau),
\end{equation}

\noindent
where $N$ is the number of evaluated images, $\mathrm{NME}_i$ is the NME for image $i$, and $1(\cdot)$ is the indicator function equal to 1 when the condition is satisfied and 0 otherwise. We then computed the normalised area under the CED curve:

\begin{equation}
\mathrm{AUC}_{\tau_0}
=
\frac{1}{\tau_0}
\int_0^{\tau_0}
\mathrm{CED}(\tau)\, d\tau ,
\end{equation}

\noindent
which normalises the area by the maximum possible value $\tau_0$, such that a perfect model yields $\mathrm{AUC}_{\tau_0}=1$. We reported AUC as a percentage at the commonly used thresholds $\mathrm{AUC}_{0.08}$ and $\mathrm{AUC}_{0.10}$.

\section{Results}

\subsection{Models results}

The evaluation results of NeoLoc-68, before fine-tuning, on public datasets, compared against selected facial landmark detectors, are summarised in Table~\ref{tab:landmarks_results_publicset}. Table~\ref{tab:nicu_models} shows the performance of several selected models alongside our own trained end-to-end NeoLoc-68 model prior to fine-tuning and after fine-tuning. Note that the public test set in these tables refers to the combined test sets of all 11 selected public datasets.

\begin{sidewaystable}[htbp]
\centering

\caption{Performance of the evaluated landmark detection models and the NeoLoc-68 model before fine-tuning on 11 public datasets. The best-performing results on the public test sets for each metric are highlighted in bold.}
\label{tab:landmarks_results_publicset}

\begin{threeparttable}
\scriptsize
\setlength{\tabcolsep}{12pt}
\renewcommand{\arraystretch}{1.4}

\begin{tabular}{@{}llrrrrrrrrrrrr@{}}
\toprule
\multicolumn{2}{@{}l}{} &
\multicolumn{11}{c}{\textbf{Datasets}} &
\multicolumn{1}{c}{\textbf{Public}} \\
\cmidrule(lr){3-13}
\textbf{Model} & \textbf{Metric} &
\textbf{XM2VTS} & \textbf{MultiPIE} & \textbf{FRGC} & \textbf{LFPW} &
\textbf{HELEN} & \textbf{AFW} & \textbf{300-W-C} & \textbf{300-W-Ch} &
\textbf{COFW} & \textbf{Menpo2D} & \textbf{WFLW} &
\textbf{test set} \\
\midrule
\midrule

\multirow{5}{*}{\textbf{Dlib} \cite{dlib09}}
& NME ($\downarrow$)        & 4.94 & 9.48 & 4.22 & 4.93 & 4.68 & 6.32 & 9.49 & 10.98 & 7.63 & 7.85 & 9.31 & 7.29 \\
& FR (\%, $\downarrow$)     & 4.67 & 36.82 & 0.62 & 4.71 & 3.51 & 16.13 & 45.71 & 54.55 & 29.21 & 33.21 & 45.29 & 25.68 \\
& DFR (\%, $\downarrow$)    & 0.21 & 14.43 & 0.00 & 4.50 & 2.15 & 8.82 & 12.50 & 18.52 & 12.75 & 11.09 & 19.62 & 10.09 \\
& AUC$_{0.08}$ ($\uparrow$) & 41.32 & 23.63 & 47.43 & 40.86 & 53.26 & 28.11 & 13.93 & 15.30 & 20.14 & 21.67 & 12.72 & 28.73 \\
& AUC$_{0.1}$ ($\uparrow$)  & 52.44 & 32.51 & 57.88 & 52.14 & 62.02 & 40.60 & 24.23 & 22.45 & 31.69 & 32.24 & 23.36 & 38.95 \\
\midrule

\multirow{5}{*}{\textbf{FAN} \cite{bulat2017far}}
& NME ($\downarrow$)        & 6.10 & 11.31 & 6.94 & 8.82 & 8.17 & 11.05 & 11.42 & 13.69 & 9.62 & 10.18 & 11.32 & 9.73 \\
& FR (\%, $\downarrow$)     & 8.69 & 66.79 & 14.74 & 56.50 & 44.88 & 91.18 & 88.24 & 96.30 & 51.96 & 70.78 & 71.96 & 56.80 \\
& DFR (\%, $\downarrow$)    & 0.00 & 0.08 & 0.00 & 0.00 & 18.24 & 0.00 & 0.83 & 0.00 & 0.00 & 4.38 & 1.73 & 2.92 \\
& AUC$_{0.08}$ ($\uparrow$) & 26.93 & 6.46 & 15.58 & 6.57 & 9.80 & 0.96 & 1.28 & 0.28 & 5.80 & 4.51 & 4.33 & 8.14 \\
& AUC$_{0.1}$ ($\uparrow$)  & 40.44 & 13.72 & 31.05 & 17.02 & 22.36 & 5.84 & 6.04 & 2.81 & 16.60 & 12.35 & 11.59 & 17.49 \\
\midrule

\multirow{5}{*}{\textbf{PIPNet} \cite{JLS21}}
& NME ($\downarrow$)        & 4.90 & 9.42 & 4.68 & 6.42 & 5.75 & 7.76 & 7.66 & 9.40 & 5.79 & 7.55 & 5.40 & 6.98 \\
& FR (\%, $\downarrow$)     & 2.54 & 53.72 & 0.00 & 14.60 & 7.65 & 42.86 & 30.95 & 66.67 & 9.09 & 33.32 & 11.38 & 26.15 \\
& DFR (\%, $\downarrow$)    & 0.00 & 0.08 & 1.96 & 31.50 & 29.83 & 79.41 & 65.00 & 55.56 & 13.73 & 21.27 & 65.51 & 22.26 \\
& AUC$_{0.08}$ ($\uparrow$) & 39.16 & 11.93 & 41.49 & 21.91 & 29.20 & 12.60 & 12.99 & 4.84 & 28.81 & 16.53 & 37.04 & 23.95 \\
& AUC$_{0.1}$ ($\uparrow$)  & 51.08 & 20.89 & 53.19 & 36.10 & 42.73 & 24.05 & 26.89 & 13.13 & 42.17 & 28.57 & 47.90 & 35.28 \\
\midrule

\multirow{5}{*}{\textbf{SPIGA} \cite{SPIGA_model}}
& NME ($\downarrow$)        & 3.94 & 7.70 & 3.29 & 4.17 & 3.19 & 4.93 & 5.95 & 8.24 & 5.42 & 5.67 & 8.75 & 5.88 \\
& FR (\%, $\downarrow$)     & 0.42 & 31.21 & 0.00 & 1.00 & 0.00 & 1.47 & 13.33 & 59.26 & 6.86 & 11.59 & 36.13 & 15.60 \\
& DFR (\%, $\downarrow$)    & --    & --    & --    & --    & --    & --    & --    & --    & --    & --    & --    & --    \\
& AUC$_{0.08}$ ($\uparrow$) & 50.85 & 27.30 & 58.93 & 48.03 & 60.06 & 38.68 & 29.17 & 10.27 & 33.12 & 35.62 & 15.62 & 36.91 \\
& AUC$_{0.1}$ ($\uparrow$)  & 60.63 & 36.46 & 67.14 & 58.34 & 68.05 & 50.70 & 41.62 & 22.15 & 45.86 & 47.04 & 26.94 & 47.15 \\
\midrule

\multirow{5}{*}{\textbf{HRNet} \cite{wang2020deep}}
& NME ($\downarrow$)        & 5.61 & 9.02 & 5.55 & 5.70 & 4.91 & 6.30 & 8.20 & 11.76 & 7.02 & 7.95 & 9.65 & 7.66 \\
& FR (\%, $\downarrow$)     & 2.33 & 36.75 & 0.41 & 4.50 & 0.64 & 10.29 & 41.67 & 77.78 & 26.47 & 34.61 & 44.25 & 26.69 \\
& DFR (\%, $\downarrow$)    & --    & --    & --    & --    & --    & --    & --    & --    & --    & --    & --    & --    \\
& AUC$_{0.08}$ ($\uparrow$) & 30.04 & 19.33 & 30.64 & 29.39 & 38.76 & 22.62 & 10.23 & 3.70 & 19.95 & 16.41 & 12.11 & 20.84 \\
& AUC$_{0.1}$ ($\uparrow$)  & 43.89 & 29.23 & 44.51 & 43.09 & 50.94 & 37.13 & 22.68 & 8.23 & 32.41 & 28.28 & 22.88 & 32.67 \\
\midrule

\multirow{5}{*}{\textbf{NeoLoc-68}\tnote{*}}
& NME ($\downarrow$)        & 3.10 & 6.13 & 2.70 & 5.40 & 4.88 & 7.11 & 6.92 & 8.72 & 6.20 & 5.51 & 7.26 & \textbf{5.37} \\
& FR (\%, $\downarrow$)     & 0.21 & 21.56 & 0.00 & 6.00 & 3.00 & 27.94 & 27.50 & 59.26 & 13.73 & 9.42 & 26.10 & \textbf{12.50} \\
& DFR (\%, $\downarrow$)    & 0.00 & 0.00 & 0.00 & 0.00 & 0.00 & 0.00 & 0.00 & 0.00 & 0.00 & 0.00 & 0.36 & \textbf{0.06} \\
& AUC$_{0.08}$ ($\uparrow$) & 61.26 & 30.24 & 66.29 & 33.48 & 39.38 & 18.08 & 19.57 & 8.88 & 28.38 & 35.59 & 22.60 & \textbf{38.00} \\
& AUC$_{0.1}$ ($\uparrow$)  & 68.97 & 41.15 & 73.03 & 46.14 & 51.26 & 30.42 & 32.55 & 19.30 & 41.32 & 47.30 & 34.24 & \textbf{48.70} \\
\bottomrule
\end{tabular}

\begin{tablenotes}[flushleft]
\footnotesize
\item \noindent
Normalised Mean Error (NME, $\downarrow$), Failure Rate (FR, $\downarrow$), Detection Failure Rate (DFR, $\downarrow$), and Area Under the Cumulative Error Curve (AUC, $\uparrow$) are reported at error thresholds 0.08 and 0.10. Lower values ($\downarrow$) indicate better performance for NME, FR and DFR; higher values ($\uparrow$) indicate better performance for AUC. 300-W-C and 300-W-Ch denote the common and challenging (iBUG) splits of 300-W, respectively.
\item[*] NeoLoc-68 model before fine-tuning.
\end{tablenotes}

\end{threeparttable}
\end{sidewaystable}

\begin{sidewaystable}[p]
\centering
\caption{Performance of the evaluated landmark detection models and NeoLoc-68 on the clinical neonatal test set. The best-performing results for each metric are highlighted in bold.}
\label{tab:nicu_models}

\begin{threeparttable}
\footnotesize
\setlength{\tabcolsep}{20pt}
\renewcommand{\arraystretch}{1}

\begin{tabular}{@{}llr@{}}
\toprule
\textbf{Model} & \textbf{Metric} & \multicolumn{1}{c}{\textbf{Clinical neonatal test set}} \\
\midrule
\midrule

\multirow{5}{*}{\textbf{Dlib} \cite{dlib09}}
& NME ($\downarrow$)        & 81.16 \\
& FR (\%, $\downarrow$)     & 76.06 \\
& DFR (\%, $\downarrow$)    & 87.46 \\
& AUC$_{0.08}$ ($\uparrow$) & 4.41  \\
& AUC$_{0.1}$ ($\uparrow$)  & 9.84  \\
\midrule

\multirow{5}{*}{\textbf{FAN} \cite{bulat2017far}}
& NME ($\downarrow$)        & 14.91 \\
& FR (\%, $\downarrow$)     & 85.11 \\
& DFR (\%, $\downarrow$)    & 25.27 \\
& AUC$_{0.08}$ ($\uparrow$) & 1.80  \\
& AUC$_{0.1}$ ($\uparrow$)  & 6.44  \\
\midrule

\multirow{5}{*}{\textbf{PIPNet} \cite{JLS21}}
& NME ($\downarrow$)        & 8.26  \\
& FR (\%, $\downarrow$)     & 39.87 \\
& DFR (\%, $\downarrow$)    & 18.02 \\
& AUC$_{0.08}$ ($\uparrow$) & 12.39 \\
& AUC$_{0.1}$ ($\uparrow$)  & 23.97 \\
\midrule

\multirow{5}{*}{\textbf{SPIGA} \cite{SPIGA_model}}
& NME ($\downarrow$)        & 13.74 \\
& FR (\%, $\downarrow$)     & 51.59 \\
& DFR (\%, $\downarrow$)    & --    \\
& AUC$_{0.08}$ ($\uparrow$) & 9.48  \\
& AUC$_{0.1}$ ($\uparrow$)  & 18.76 \\
\midrule

\multirow{5}{*}{\textbf{HRNet} \cite{wang2020deep}}
& NME ($\downarrow$)        & 15.60 \\
& FR (\%, $\downarrow$)     & 70.14 \\
& DFR (\%, $\downarrow$)    & --    \\
& AUC$_{0.08}$ ($\uparrow$) & 5.06  \\
& AUC$_{0.1}$ ($\uparrow$)  & 11.45 \\
\midrule

\multirow{5}{*}{\textbf{NeoLoc-68} \textit{(Before fine-tuning)}}
& NME ($\downarrow$)        & 9.34  \\
& FR (\%, $\downarrow$)     & 55.78 \\
& DFR (\%, $\downarrow$)    & 5.30  \\
& AUC$_{0.08}$ ($\uparrow$) & 8.40  \\
& AUC$_{0.1}$ ($\uparrow$)  & 18.18 \\
\midrule

\multirow{5}{*}{\textbf{NeoLoc-68} \textit{(After fine-tuning)}}
& NME ($\downarrow$)        & \textbf{6.36}  \\
& FR (\%, $\downarrow$)     & \textbf{22.30} \\
& DFR (\%, $\downarrow$)    & \textbf{1.77}  \\
& AUC$_{0.08}$ ($\uparrow$) & \textbf{29.24} \\
& AUC$_{0.1}$ ($\uparrow$)  & \textbf{40.25} \\
\bottomrule
\end{tabular}

\begin{tablenotes}[flushleft]
\footnotesize
\item \noindent
Normalised Mean Error (NME, $\downarrow$), Failure Rate (FR, $\downarrow$), Detection Failure Rate (DFR, $\downarrow$), and Area Under the Cumulative Error Curve (AUC, $\uparrow$) are reported at error thresholds 0.08 and 0.10. Lower values ($\downarrow$) indicate better performance for NME, FR and DFR; higher values ($\uparrow$) indicate better performance for AUC.
\end{tablenotes}

\end{threeparttable}
\end{sidewaystable}

\section{Discussion}

\subsection{Results on public datasets}
NeoLoc-68, before fine-tuning, achieves the best performance across all evaluation metrics on the public test set, outperforming all other evaluated models as shown in Table \ref{tab:landmarks_results_publicset}. However, when analysing the results on individual datasets, the ranking varies. On the 300-W dataset, NeoLoc-68 achieves the second-best NME, slightly behind SPIGA. SPIGA operates using the ground-truth face location during testing rather than performing face detection, providing an advantage over end-to-end models that must first localise the face before landmark prediction. In addition, SPIGA was originally trained on the 300-W dataset, which may further contribute to its performance on this benchmark. Because the original training–testing splits used by previous works are not always publicly available, and given the preprocessing applied in this study, including single-face image selection and the dataset splitting strategy (sections~\ref{sec:data_preprocessing} and \ref{sec:model_training}), it is not possible to determine which images were part of SPIGA’s original training or evaluation phases. This limitation extends to other evaluated models trained on the same datasets. Consequently, some images included in our test set may have been seen during training in previously published works, which may have favoured their performance in this comparison.

NeoLoc-68 outperforms all other models on the public test set. We trained the model on samples drawn from all available datasets, allowing it to capture domain-specific characteristics unique to each dataset. This highlights the importance not only of the quantity of training data, but also the critical role that dataset diversity plays in developing generalisable facial landmark detection models.

NeoLoc-68 achieved a near-zero DFR across all datasets except WFLW, which recorded a value of 0.36\%. This may have been achieved due to the weight initialisation method we employed. By transferring pre-trained weights from our face detector, which was trained on millions of face images, we provided the model with a strong prior for facial structure and localisation. This level of consistency is not matched by any other evaluated model, reinforcing the robustness of our end-to-end approach.

\subsection{Results on the neonatal research dataset}

\subsubsection{Before fine-tuning}

Results shown in Table \ref{tab:nicu_models} highlight the challenges presented by the clinical neonatal domain when compared to publicly available datasets. Across all models, a clear decrease in performance is evident when comparing the clinical neonatal results to those obtained on the public test set presented in Table \ref{tab:landmarks_results_publicset}. For instance, the NME for Dlib increases from 7.29 to 81.16, and DFR rises from 10.09\% to 87.46\%.

HRNet, when evaluated using cropped faces, achieves $\mathrm{AUC}_{0.08}$ and $\mathrm{AUC}_{0.10}$ of 5.06 and 11.45 respectively, with a high FR at 70.14\%. A similar behaviour was observed with SPIGA, which despite being provided face locations, still recorded a high FR of 51.59\%. This suggests that local appearance, rather than face detection alone, may be the primary bottleneck for performance in neonatal clinical environments.

FR consistently presents a challenge across all models; even the best-performing model, PIPNet, misses nearly 40\% of frames, highlighting the domain gap between adults and neonates in clinical conditions characterised by unconstrained head poses and frequent occlusions from clinical intervention or medical equipment.

PIPNet achieved the lowest landmark localisation error, with an NME of 8.26, followed by NeoLoc-68 with 9.34 and then SPIGA with 13.74. However, NeoLoc-68 outperformed all other models in terms of DFR, achieving a robust result of 5.30\%, markedly better than the second-best model, PIPNet at 18.02\%, and the third-best, FAN at 25.27\%. This favours NeoLoc-68 particularly in critical environments such as the neonatal unit, as its NME of 9.34 is calculated from 94.70\% of the test set, whereas PIPNet’s NME of 8.26 is based on 81.98\% of the test set due to higher detection failures. We propose two possible approaches for evaluation: (1) In end-to-end settings where the face location is not known in advance (e.g., video monitoring of neonates on full frames), detection reliability is critical, and a weighted NME metric could be introduced to better reflect model performance; (2) In less detection-critical scenarios, such as controlled benchmarking studies where the face location is predefined or provided as ground truth, it may be acceptable to follow existing practices where the face location is always provided. The second approach is not suitable in our scenario, as our goal is to introduce an end-to-end solution capable not only of localising facial keypoints but also of processing an entire frame directly, without relying on cropping or using ground truth face bounding boxes. Such an end-to-end model is more beneficial for real-world implementations.

The results of NeoLoc-68, with the trade-off between detection and NME score, suggest that our backbone and data-augmentation strategy already provide strong generalisation capabilities to neonatal images before any fine-tuning specific to neonatal conditions. However, final landmark refinement still benefits from domain-specific cues that are absent in public data. Before any neonatal fine-tuning, NeoLoc-68 delivers state-of-the-art detection reliability and competitive localisation accuracy, outperforming all baselines on DFR and ranking third on three other metrics. Nonetheless, the 55.78\% failure rate indicates that there is much room for improvement.

\subsubsection{After fine-tuning}

Table \ref{tab:nicu_models} shows the performance of NeoLoc-68 after fine-tuning on the neonatal research dataset. We observed clear and consistent improvements across all evaluation metrics. The NME decreased from 9.34 to 6.36, alongside an improvement in DFR to 1.77\%. This indicates more images contributed to NME calculation, yet performance still improved with NME decreasing by 31.91\%, demonstrating substantial gains in localisation and detection accuracy.

The FR dropped notably from 55.78\% to 22.30\%, reflecting the model's improved ability to reliably predict valid landmarks across challenging frames. Additionally, both AUC scores improved as $\mathrm{AUC}_{0.08}$ rose from 8.40 to 29.24, and $\mathrm{AUC}_{0.10}$ from 18.18 to 40.25. These results highlight a shift in the overall distribution of localisation errors, with a much higher proportion of predictions falling within acceptable error thresholds.

These results demonstrate the benefit of domain-specific fine-tuning when adapting facial landmark models to neonatal clinical environments. While NeoLoc-68 already showed promising generalisation prior to adaptation, especially in terms of low detection failure rate, fine-tuning allows it to more effectively handle the unique characteristics of neonatal images, such as different facial proportions, frequent occlusions, and unconstrained head poses.

Two main factors might contribute to further improvements in FR and AUC metrics. Firstly, the availability of labelled clinical neonatal data remains limited, particularly regarding the number of participants and diversity across clinical settings. Secondly, the absence of clear guidelines during the annotation process also affects the model’s performance.

\subsection{Reliance on external face detector}
Our proposed model offers end-to-end facial landmark detection, eliminating the need for separate face detection or manual cropping steps at inference time, allowing it to process full images directly. During data preprocessing, cropping was applied only to multi-face images to isolate the annotated face, while images containing a single face were used in their original full-image form.

Several studies report training on cropped face images \cite{zhu2019robust,wang2020deep,SPIGA_model}. While cropping can accelerate training by reducing input size and complexity, it introduces an implicit bias: the model learns that faces are centred and occupy most of the image area. Most existing models require an external face detector or rely on cropping the face region during preprocessing at inference time. Such assumptions limit the model’s ability to generalise to unconstrained, real-world settings where faces may appear at varied positions and scales.

\subsection{Evaluation metrics}
When applying the NME metric, the choice of normalisation factor $d_i$ in Eq~(\ref{eq:NME}) varies across studies. Common options include the distance between the inner eye corners, the outer eye corners, the pupil centres, or the diagonal of the ground truth bounding box. This choice directly influences the reported NME values, as it determines how localisation errors are scaled. The smaller the normalisation factor $d_i$, the larger the resulting NME for a given prediction error. Some studies adopt more lenient normalisation strategies, such as using outer eye corners or bounding box diagonals, which tend to yield lower NME values.

We used the inter-ocular distance based on the Euclidean distance between the centre of the eyes, computed as the mean of the six eye landmarks for each eye. This results in a shorter normalisation factor compared to outer-eye-corner distance, and thus yields higher NME values for the same level of prediction error. This partially explains the higher NME values reported in our results compared to some published benchmarks.

We chose to avoid bounding-box-based or dataset-specific normalisation. Our dataset is composed of images drawn from multiple sources (e.g., 300-W, iBUG, COFW, WFLW), and using different normalisation factors $d_i$ across subsets would compromise the consistency and fairness of the evaluation, as identical prediction errors would contribute differently to the overall NME depending on the dataset from which each image originated. We also aimed to apply a uniform metric across all datasets and models. Although this choice may affect the absolute NME values, it ensures that comparisons remain unbiased and globally consistent.

\subsection{Limitations}
A key limitation in facial landmark detection, both in neonatal clinical settings and more generally when using the standard 68-point iBUG markup, is the lack of formalised guidelines for point placement. There is no universally agreed-upon method for determining the location of each landmark. For example, as illustrated in Fig~\ref{fig:68_markup}(a), it is unclear where Point 1 of the jawline should be placed, or how the jawline points should be distributed. This absence of published annotation guidelines introduces inconsistencies across datasets and also affects the manual annotation process in our work. In the neonatal clinical setting specifically, a further challenge is the limited visibility of certain facial features. In some infants, features such as the eyebrows are faint or difficult to distinguish, which complicates annotation and may affect model performance.

Developing clear, standardised point-placement guidelines would benefit the research community and improve both dataset quality and model performance. One possible solution is to define landmark locations relative to anatomical facial references. For example, the beginning of the jawline could be anchored relative to the top edge of the subject’s right ear. Similarly, the nose bridge (Point 28 in Fig~\ref{fig:68_markup}(a)) could be placed at the midpoint of a horizontal line connecting the inner corners of the eyes (Points 40 and 43). Additionally, scenarios involving self-occlusion or when parts of the face fall outside the frame remain unresolved and require clearer conventions.

Manually annotating large datasets with 68 facial landmarks is both expensive and time-consuming. As a result, many publicly available datasets have been annotated using semi-supervised methods. A major concern with these approaches is that the accuracy of point placement is often not rigorously validated (see Supplementary Material S5).

Upon running inference on clinical neonatal examples, the model tends to struggle with frames extracted from videos taken from a side view, where most of the face is not frontal. This might be improved by increasing the rotation augmentation beyond our current selection (within $[-30^\circ, +30^\circ]$), or by rotating the frames before inference. We also observed that the model frequently predicted partially closed eyes, even when the infant’s eyes were open, suggesting a bias towards closed-eye appearances. We investigated this by computing the Eye Aspect Ratio (EAR) across our neonatal research dataset, which showed that eye openness is mostly near zero (see Supplementary Material S6), indicating a strong skew towards closed eyes in the data. This likely affected the model’s ability to learn eye status. Mitigations include rebalancing the dataset with more open-eye images or increasing the loss weight for the eye landmarks.

A final limitation of our work is the absence of inter-annotator agreement analysis. All annotations were performed manually by one of the authors, and the time-intensive nature of the task made it impractical to involve additional annotators. Nevertheless, the absence of such a comparison means that the reliability of the annotations could not be quantitatively assessed.

\section{Conclusion}

We presented a complete end-to-end 68-point facial landmark detection pipeline tailored for neonatal clinical environments. The pipeline begins with the construction of a custom neonatal research dataset, created from video recordings and manually annotated with 68 facial landmarks following the iBUG standard. We described the dataset preprocessing, annotation, training and evaluation processes for both public and neonatal research datasets to ensure consistency and reproducibility.

We proposed NeoLoc-68, a keypoint detection model designed to predict 68 facial landmarks, leveraging pre-trained weights from a face detection model trained on millions of facial images. We demonstrated the model’s capacity to generalise across domains, achieving promising results even before fine-tuning on neonatal data.

Despite these achievements, several challenges persist. The clinical neonatal domain introduces difficulties, including limited data availability, poor visibility of facial features due to clinical interventions, and high variability in head poses. Manual annotation of 68 landmarks is especially time-consuming and resource-intensive. To address this, future work could incorporate semi-automated annotation techniques, similar to our approach in selecting single-face images. For example, multiple models could be used to annotate the same frame, with their outputs averaged. This would not only reduce the manual burden but also serve as a starting point for annotating sequential frames, paving the way for integrating temporal information into model training. Another promising direction is the use of 3D-trained models, which may offer improved robustness to occlusion, particularly self-occlusion. Incorporating inter-annotator agreement assessments and establishing clearer annotation guidelines could further enhance data quality and model reliability.

Overall, this work lays the foundation for neonatal facial landmark detection in real-world clinical settings. With continued development, particularly in data expansion and methodological refinement, this pipeline can evolve into a practical tool for neonatal monitoring and pain assessment in clinical environments, and serve as a starting point for facial pain-related feature analysis.

\section*{Ethics statement}

The neonatal research data were collected under approvals from the London-South East Research Ethics Committee for the Petal trial (21/LO/0523) and the South Central - Oxford C Research Ethics Committee for “Investigating Pain in the Developing Human Brain” (12/SC/0447). Written informed parental consent was obtained for all neonates prior to participating, and the studies conformed to the Declaration of Helsinki and Good Clinical Practice standards.

\section*{Funding}

A.B. acknowledges support from a scholarship awarded by King Saud University.
M.V. was funded by the Podium Institute for Sports Medicine and Technology, University of Oxford.

\section*{CRediT authorship contribution statement}
M.C. and R.S. collected the data. R.S., L.T. and M.V. conceptualised the experiment and supervised the project. A.B. conducted the experiments, performed the data analysis and wrote the first draft of the manuscript. All authors reviewed the manuscript.

\section*{Declaration of competing interest}

The authors declare no competing interests.

\section*{Data availability}

The neonatal research dataset used in this study is not publicly available due to its sensitive nature, terms of parental consent, and restrictions imposed by the approved ethics protocol to protect the privacy of the infants involved in the study.\\

\noindent
The public datasets used are available from their respective original sources and are subject to the terms and conditions of those dataset providers.

\section*{Code availability}
Annotations generated from the public datasets, training instructions and the corresponding code will be made available at: https://github.com/lcmtlab/InfantFace upon acceptance.

\section*{Acknowledgements}
We thank the parents and infants who participated in the original studies from which the neonatal research dataset was derived. The views expressed in this publication are those of the authors and not necessarily those of the NIHR, the University of Oxford, National Health Service, or the UK Department of Health and Social Care.

\newpage
\bibliographystyle{unsrtnat}
\bibliography{NeoLoc_refs}

\clearpage
\section*{Supplementary Material}

\setcounter{figure}{0}
\setcounter{table}{0}
\renewcommand{\thefigure}{S\arabic{figure}}
\renewcommand{\thetable}{S\arabic{table}}

\maketitle
\thispagestyle{firstpage}

\newpage

\section*{Supplementary Material S1: Standardisation of facial landmark annotations to the 68-point iBUG markup}
\label{S1_Fig}

\begin{figure}[h]
	\centering
	\includegraphics[width=\textwidth]{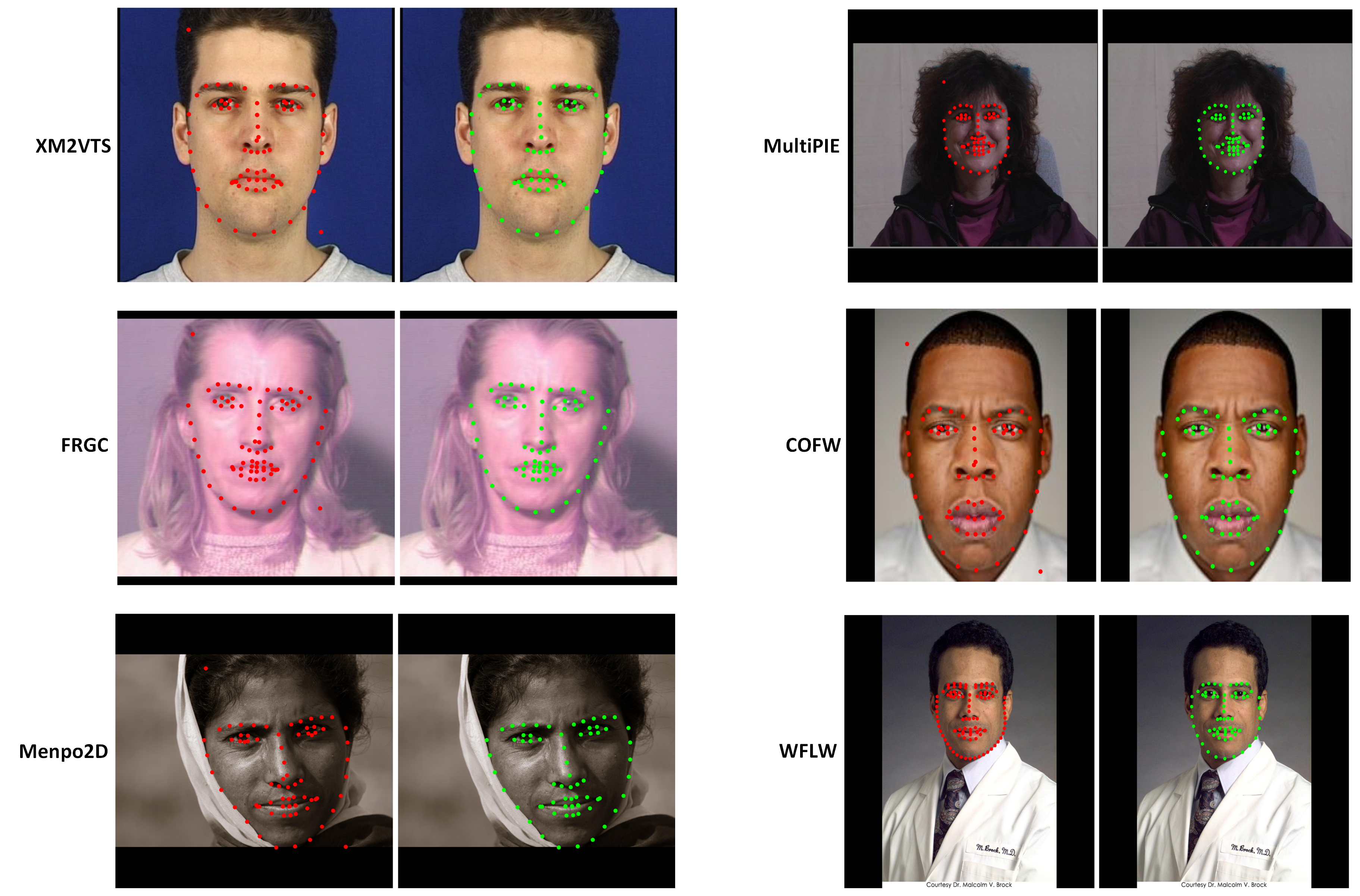}
\caption{
\textbf{Examples of landmark standardisation across selected datasets}. Different face datasets provide annotations with different numbers and arrangements of landmarks. To obtain a consistent representation for subsequent analysis, all annotations were converted to the standard 68-landmark iBUG format. For each dataset, the original landmark annotations are shown in red, and the resulting standardised 68-landmark annotations are shown in green. For XM2VTS, MultiPIE, FRGC, COFW and Menpo2D, this involved removing non-standard landmarks such as bounding box corner points, duplicate nose tip or mouth edge landmarks, and eye pupil annotations. For WFLW, the conversion involved retaining every other jawline landmark, removing the additional lower eyebrow landmarks, and discarding the pupil and extra eyelid landmarks. These examples illustrate how heterogeneous landmark annotations were harmonised into a unified 68-landmark configuration. Some images have been cropped and padded for visual alignment.
}
\end{figure}

\newpage

\section*{Supplementary Material S2: Summary of the pre-trained face detector used for initialisation}

We initialised our landmark detector from a YOLOv11m-based face detector (“InfantFace”) developed for neonatal clinical environments. The model was first trained on large-scale public datasets (VGGFace2, CelebA, FDDB and WIDER FACE), then domain-adapted on clinical neonatal videos recorded during routine procedures. After adaptation, the model achieved an $\text{AP}_{50}$ of 0.96 ($\text{AP}_{50:95}$ of 0.67), indicating sufficiently accurate localisation to act as a facial prior for landmarks. We use these weights solely as generic facial priors; all landmark learning is performed on the neonatal landmark datasets described in the Methods. Full details are provided in Bin-Obaid et al. [S1].

\newpage

\section*{Supplementary Material S3: Computing Hue Saturation Value (HSV) for augmentation }

\label{S2_Appendix}

We converted all images from RGB to Hue Saturation Value (HSV) colour space. For each image $i$, we computed the mean of each channel $(\bar H_i,\bar S_i,\bar V_i)$ using OpenCV’s scales. We then computed the standard deviation across images of these per-image means to obtain $\sigma_H$, $\sigma_S$ and $\sigma_V$. To set YOLO’s HSV augmentation amplitudes, each $\sigma$ was multiplied by 3 (capturing roughly $\pm3\sigma$ of the colour distribution) and normalised by the channel’s numeric range, then clipped to $[0,1]$.

\newpage

\section*{Supplementary Material S4: Face detection loss function}
\label{S3_Appendix}

The total loss is a weighted sum of three components: bounding box loss (\(\mathcal{L}_{\mathrm{box}}\)), classification loss (\(\mathcal{L}_{\mathrm{cls}}\)) and Distribution Focal Loss (DFL) (\(\mathcal{L}_{\mathrm{dfl}}\)), given by:

\begin{equation}
	\begin{aligned}
		\mathcal{L}_{\mathrm{total}}
		&=
		\lambda_{\mathrm{box}} \,\mathcal{L}_{\mathrm{box}}
		\;+\;
		\lambda_{\mathrm{cls}} \,\mathcal{L}_{\mathrm{cls}}
		\;+\;
		\lambda_{\mathrm{dfl}} \,\mathcal{L}_{\mathrm{dfl}}
	\end{aligned}
\end{equation}

\noindent

where \(\lambda_{\mathrm{box}}\), \(\lambda_{\mathrm{cls}}\) and \(\lambda_{\mathrm{dfl}}\) are the gains for the bounding box, classification and DFL losses respectively.

\noindent
Bounding box loss (\(\mathcal{L}_{\mathrm{box}}\)) penalises differences between the predicted and ground-truth bounding boxes. It is computed using the Complete-IoU (CIoU), described in [S2]. CIoU extends standard IoU by adding penalties for the normalised distance between box centres and for aspect-ratio mismatch. The \(\mathcal{L}_{\mathrm{box}}\) is defined as:

\begin{equation}
  \mathcal{L}_{\mathrm{box}}  = 1 - \mathrm{CIoU}
  \label{eq:bbox}
\end{equation}

\noindent
with
\begin{equation}
\begin{aligned}
  \mathrm{CIoU} &= \mathrm{IoU} - \frac{\rho^{2}(p,\, p_{\mathrm{gt}})}{c^{2}} - \alpha V \\[8pt]
  \alpha &= \frac{V}{(1-\mathrm{IoU}) + V} \\[8pt]
  V &= \frac{4}{\pi^{2}}\!\left(\arctan\frac{w}{h}-\arctan\frac{w_{\mathrm{gt}}}{h_{\mathrm{gt}}}\right)^{2}
\end{aligned}
\label{eq:ciou}
\end{equation}

\noindent
where \(\rho^{2}(p, p_{\mathrm{gt}})\) is the squared distance between the box centres, \(c\) is the diagonal of the smallest enclosing box that covers both predicted and ground-truth boxes and \(w,h\) and \(w_{\mathrm{gt}},h_{\mathrm{gt}}\) are the widths and heights of the predicted and ground-truth boxes, respectively. $V$ is the aspect-ratio penalty and $\alpha$ is the weight for $V$. Both $\alpha$ and $V$ increase the penalty when the IoU is high and the aspect ratio differs.\\

\noindent
The classification loss (\(\mathcal{L}_{\mathrm{cls}}\)) is defined as the binary cross-entropy for each predicted box:
\begin{equation}
	\begin{aligned}
		\mathcal{L}_{\mathrm{cls}}
		&=
-\frac{1}{S}\sum_{i=1}^{N}\Bigl[y_i\log\sigma(x_i)+(1-y_i)\log(1-\sigma(x_i))\Bigr]
\end{aligned}
	\label{eq:cls}
\end{equation}

\noindent
where \(x_i\) is the predicted logit for the \(i\)-th candidate box, \(y_i \in [0,1]\) is a target indicating how strongly prediction \(i\) matches a ground-truth face, \(\sigma(\cdot)\) is the sigmoid function, \(N\) is the total number of candidate boxes in the batch and \( S \) is the normalisation factor defined as the sum of the targets over all candidate boxes.

\noindent
The DFL loss ($\mathcal{L}_{\mathrm{dfl}}$) from [S3] is defined as:

\begin{equation}
	\mathcal{L}_{\mathrm{dfl}} = -\left[(y_{i+1} - y)\log P_i + (y - y_i)\log P_{i+1}\right]
	\label{eq:dfl_eq}
\end{equation}

\noindent
where \(y\) is the continuous regression target, \(y_i\) and \(y_{i+1}\) are the two adjacent discrete values. \(P_i\) and \(P_{i+1}\) are the predicted probabilities corresponding to \(y_i\) and \(y_{i+1}\), respectively.

\newpage

\section*{Supplementary Material S5: Example of incorrect landmark annotations from a public dataset}
\label{S2_Fig}

\begin{figure}[h]
	\centering
	\includegraphics[width=\textwidth]{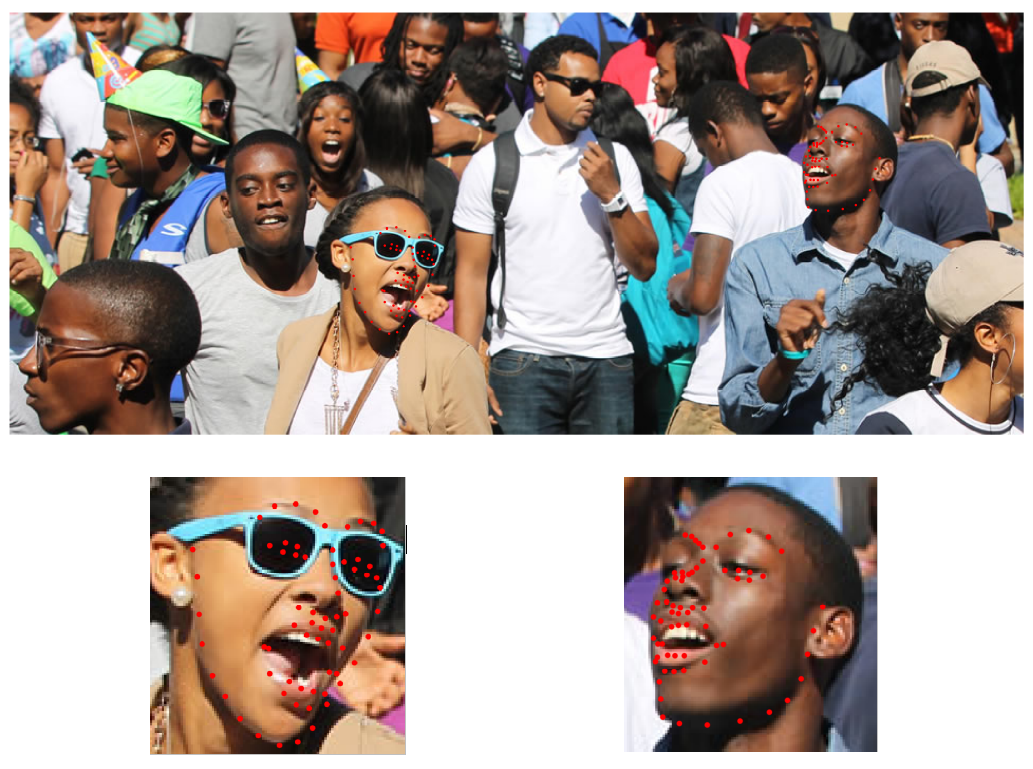}
	\caption
	{
		\textbf{Example of incorrect landmark annotations from a public dataset}. Only two faces are annotated (keypoints shown in red) in the upper image. The landmark points shown in red are the original annotations distributed with the public dataset and were not produced as part of this study. The lower images show close-ups of the annotated faces, where misalignments in landmark placement can be observed. In the left image, landmarks around the eyes are placed directly on the sunglasses, with no visible justification for eye shape behind the occlusion. In both images, the mouth annotations are misaligned: in the right image, landmarks for the upper lip are placed on the teeth, and points for the lower lip's upper edge fall in the middle of the lip rather than along its boundary.
	}
\end{figure}

\newpage

\section*{Supplementary Material S6: Distribution of the Eye Aspect Ratio (EAR)}
\label{S3_Fig}

\begin{figure}[h]
	\centering
	\includegraphics[width=.9\textwidth]{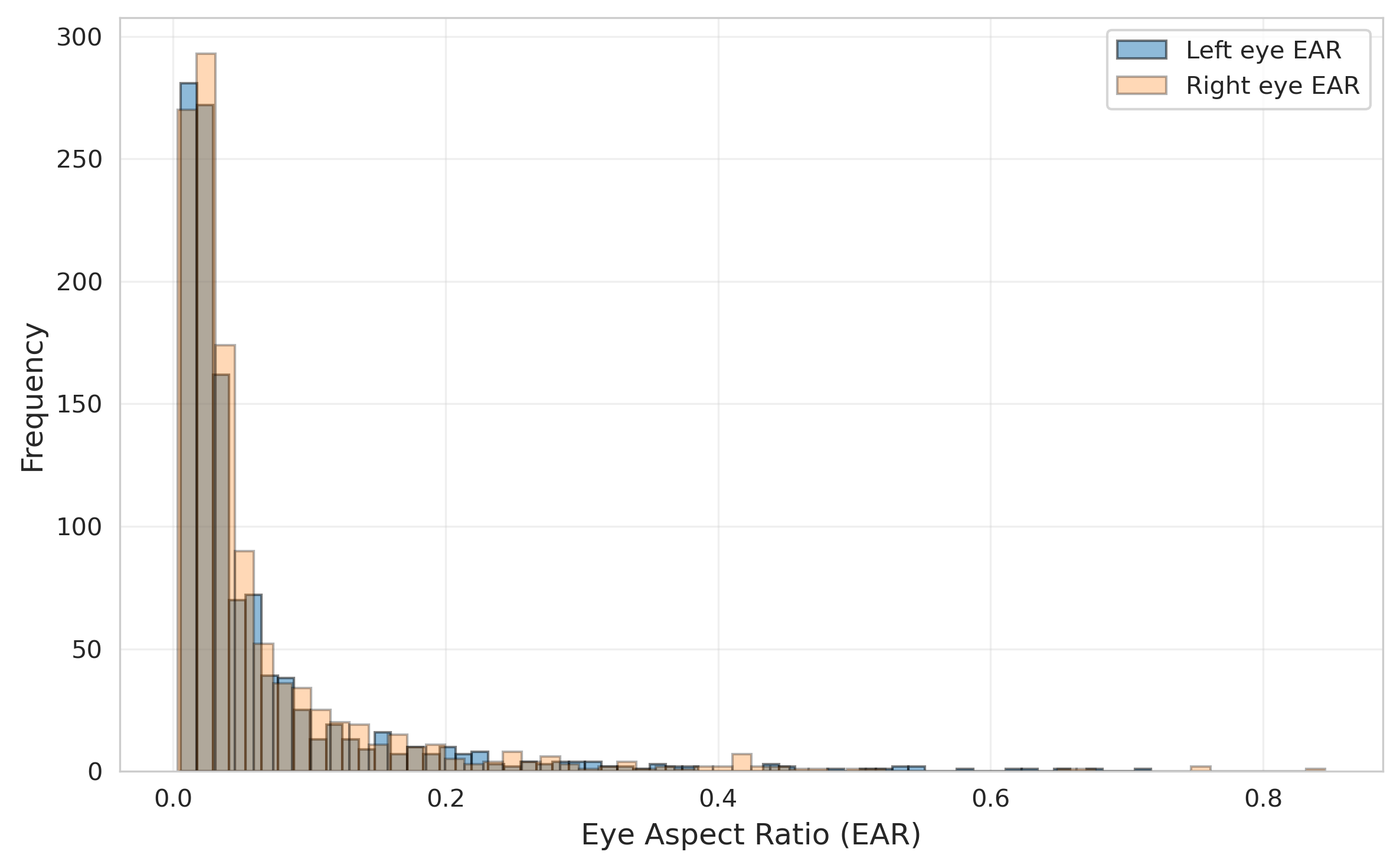}
	\caption
	{
		\textbf{Distribution of the Eye Aspect Ratio (EAR) in the clinical neonatal dataset (left vs right eye)}. The histogram is right-skewed and heavily concentrated near zero, indicating that most eyes are close to closed. EAR was computed per eye using the standard definition, i.e., the average vertical lid gap divided by the inter-canthal width.
	}
\end{figure}

\newpage

\section*{Supplementary references}

\noindent
[S1] Bin-Obaid et al. “InfantFace: Detecting infant faces in neonatal clinical environments”.

\vspace{5mm}

\noindent
[S2] Zheng Z et al. Enhancing geometric factors in model learning and inference for object detection and instance segmentation. IEEE transactions on cybernetics. 2021 Aug 26;52(8):8574-86.

\vspace{5mm}

\noindent
[S3] Li X et al. Generalized focal loss: Towards efficient representation learning for dense object detection. IEEE transactions on pattern analysis and machine intelligence. 2022 Jun 9;45(3):3139-53.

\end{document}